\pdfoutput=1
\documentclass{article}

\usepackage[preprint,nonatbib]{nips_2018}

\usepackage[utf8]{inputenc} % allow utf-8 input
\usepackage[T1]{fontenc}    % use 8-bit T1 fonts
\usepackage{hyperref}       % hyperlinks
\usepackage{url}            % simple URL typesetting
\usepackage{booktabs}       % professional-quality tables
\usepackage{amsfonts}       % blackboard math symbols
\usepackage{nicefrac}       % compact symbols for 1/2, etc.
\usepackage{microtype}      % microtypography
\usepackage{amsmath}        % Math package I added
\usepackage{graphicx}
\usepackage[numbers]{natbib}  % "authoryear" format causes issues on arxiv.

\title{Twin-GAN -- Unpaired Cross-Domain Image Translation with Weight-Sharing GANs}

\author{
  Jerry Li \\
  Google\\
  1600 Amphitheatre Parkway, Mountain View, CA 94040 \\
  \texttt{jerrylijiaming@google.com} \\
}

\begin{document}
% \nipsfinalcopy is no longer used

\maketitle

\begin{abstract}
  We present a framework for translating unlabeled images from one domain into analog images in another domain. We employ a progressively growing skip-connected encoder-generator structure and train it with a GAN loss for realistic output, a cycle consistency loss for maintaining same-domain translation identity, and a semantic consistency loss that encourages the network to keep the input semantic features in the output. We apply our framework on the task of translating face images, and show that it is capable of learning semantic mappings for face images with no supervised one-to-one image mapping.
\end{abstract}

\section{Introduction}

The ability of a human to transfer scenes of what they see to a painting has long existed since the cradle of civilization. The earliest cave paintings that we know of is not a photo-realistic representation, but rather an abstract depiction of hunting scenes. Yet when it comes to developing an algorithm that mimics such ability, we are still limited in what we can achieve.

Recent development in neural networks has greatly improved the generalization ability of image domain transfer algorithms, in which two parallel lines of development prevails: using Neural Style Transfer\cite{gatys2015neural} and using Generative Adversarial Networks(GAN)\cite{goodfellow2014generative}. 

Neural Style Transfer defines a content loss that captures the semantic information loss in transferred image, and a style loss often defined as the statistical correlation (e.g. a Gramian matrix) among the extracted features. It utilizes a trained object detection network such as VGG19\cite{simonyan2014very} for feature extraction. As a result, style transfer gives less satisfactory results when the target domain contains object representations not seen by the pretrained object detection network\cite{zhang2017style}. Furthermore, due to the content loss being defined as the L2 difference in convolutional layers of object detection network, Style Transfer based methods fails to fundamentally change the object layout when applied to domain transfer and often relies on the objects in the content and style images having similar spacial proportions in the first place. When semantic objects fails to align, e.g. body proportions in paintings are different from those in real life, style transfer leads to less desirable results.

On the other hand, GAN has shown more promising results when provided paired data belonging to the two domains of interest. Previous works such as \cite{isola2017image} and \cite{zhang2017style} all show promising results when paired datasets are available or can be generated. However such datasets are often expensive to obtain.

Current works on unpaired cross-domain image translation tasks, including but not limited to \cite{taigman2016unsupervised,yoo2016pixel,zhu2017unpaired,choi2017stargan,royer2017xgan}, usually rely on the "shared latent space" assumption. That is, given two related domains \textit{S} and \textit{T}, for item $s\in S$, there exists a unique item $t\in T$ where \textit{s} and \textit{t} shares the same latent encoding \textit{z}. However, this assumption may not always hold for all datasets and forcing such assumption on the system may result in model collapse or unrealistic-looking outputs, especially for domains that are vastly different in appearance.

To address the issues mentioned above, we introduce the Twin-GAN network and demonstrates its usage on the task of translating unpaired unlabeled face images.

\section{Related Work}

\subsection{Domain Transfer through Style Transfer}

\paragraph{Conditional Style Transfer}
By using conditional instance normalization, which learns a separate set of parameters $\gamma^s$ and $\beta^s$ in the instance normalization layers for each style $s$, \cite{dumoulin2017learned} expanded on the Neural Style Transfer methods with a single network capable of mix-and-matching 32 styles. Future works such as \cite{huang2017arbitrary} further improved on the idea by adaptively computing $\gamma^s$ and $\beta^s$ based on the input style image and proposed real-time arbitrary style transfer using a feed-forward network. 

\paragraph{Image Analogy through Patch Matching}

\cite{liao2017visual} propose a structure-preserving image analogy framework based on patch matching method. Given two structurally similar images, it extracts the semantic features using an object detection network. For each feature vector at one position, it finds a matching feature from the style image lying within the same fixed sized patch. The image is then reconstructed using the matched features from the style image. The effect is impressive, yet its application is greatly limited by its requirement on the input images -- the images must be spatially and structurally similar since features can only come from a vicinity determined by the perspective field of the network and the patch size. Nonetheless, it is powerful for some use cases and can achieve state-of-the-art results, e.g. in harmonic photoshopping of paintings\cite{luan2018deep}.

\subsection{Domain Transfer through Generative Adversarial Networks}

The Generative Adversarial Network\cite{goodfellow2014generative} frameworks synthesizes data points of a given distribution using two competing neural networks playing a minimax game. The Generator $G$ takes a noise vector sampled from a random distribution $Z$ and synthesizes data points $G(z)$ and the Discriminator $D$ tells apart real data points from the generated ones.

\paragraph{Supervised Image Translation}

\cite{isola2017image} propose a conditional GAN framework called "pix2pix" for image translation task using a paired dataset. In order to make preserving details in the output image much easier, it uses U-Net, which adds skip connection between layer $i$ in the encoder and layer $n-i$ in the decoder where $n$ is the total number of layers. \cite{isola2017image} is applied on numerous tasks such as sketch to photo, image colorization, and aerial photo to map. Similarly, \cite{zhang2017style} use a conditional GAN with Residual UNet\cite{ronneberger2015u} framework supplemented with embeddings from VGG19\cite{simonyan2014very} for sketch colorization task. Both show that conditional GAN and UNet are effective when paired data are readily available or can be generated.

\paragraph{Unsupervised Domain Transfer}

For an unsupervised domain transfer task where paired datasets are unavailable but labeled data is available for the source domain, DTN\cite{taigman2016unsupervised} transfers images in the source domain to the target domain while retaining their labeled features. It contains a pretrained feature extractor $F$ and a generator $G$ on top of $F$. The DTN is trained using a GAN loss for realistic output, a feature consistency loss for preserving key features after translation, and an identity loss for making sure the network acts as an identity mapping function when supplied with inputs sampled from the target domain. DTN's need for a pretrained feature extractor implies that the result will be limited by the quality and quantity of the labeled features.

\paragraph{Unsupervised Image Translation}
In \cite{liu2017unsupervised} the GAN-VAE based UNIT framework was introduced to handle the domain transfer task on an unpaired unlabeled dataset. It made the cycle-consistency hypothesis where for a pair of corresponding images $(x_1, x_2)$ from two domains, there exists a shared latent embedding $z$ where one-to-one mapping exists for $(x_1,z)$ and $(x_2,z)$., the GAN-VAE based UNIT framework uses this hypothesis to learn a pair of encoders mapping images from two different domains into the same latent space and a pair of generators conditioned on the latent encoding to translate the embedded image back into the two domain space. Specifically, the latent space is constrained to a Gaussian distribution by the VAE framework and weights are shared in the higher-level(deeper) layers of the encoders and decoders. It showed promising results in domain transfer tasks such as day/night scenes, canine breeds, feline breeds, and face attribute translation. 

Similarly, \cite{zhu2017unpaired} proposed the CycleGAN framework for unpaired image translation that relies on a GAN loss and a cycle consistency loss term defined as $L_{cyc}(G,F) = \mathbb{E}_{x, y} ||F(G(x)) - x||_1 + ||G(F(y)) - y||_1$.  This is similar to the shared latent embedding assumption without specifying the latent embedding distribution. The CycleGAN framework showed some success when applied to a range of classic image translation tasks. Some of its failure cases include over-recognizing objects and not being able to change the shape of the object during translation (e.g. outputting an apple-shaped orange).

Finally, \cite{choi2017stargan} introduced a multi-domain transfer framework and applied it on face synthesis using labeled facial features. \cite{royer2017xgan} trained an adversarial auto-encoder using cycle-consistency, reconstruction loss, GAN loss, and a novel teacher loss where knowledge is distilled from a pretrained teacher to the network. It showed some success on two all-frontal face dataset: the VGG-Face and the Cartoon Set.

\section{Method}
\label{method}

\subsection{Objective}
\label{objective}

Given unpaired samples from two domains $\mathbf{X_1}$ and $\mathbf{X_2}$ depicting the same underlying semantic object, we would like to learn two functions, $F_{1\rightarrow 2}$ that transforms source domain items $x_1 \in \mathbf{X_1}$ to the target domain $\mathbf{X_2}$, and $F_{2\rightarrow 1}$ that does the opposite. Following past literatures\cite{taigman2016unsupervised,zhu2017unpaired,choi2017stargan,royer2017xgan}, our TwinGAN architecture uses two encoders, $E_{1}$ and $E_{2}$, two generators $G_{1}$ and $G_{2}$, and two Discriminators $D_{1}$ and $D_{2}$ for domain $\mathbf{X_1}$ and $\mathbf{X_2}$ respectively, such that $G_{2}(E_{1}(x_1)) \in \mathbf{X_2}$ and $G_{1}(E_{2}(x_2)) \in \mathbf{X_1}$.

\subsection{Architecture}
\label{architecture}

\paragraph{Progressive GAN}

We follow the current state of the art in image generation and adapts the PGGAN structure\cite{karras2017progressive}. In order to improve the stability of GAN training, to speed up training process, and to output images with higher resolution, PGGAN progressively grows the generator and discriminator together and alternates between growing stages and reinforcement stages. During the growing stage, the input from a lower resolution is linearly combined with a higher resolution. The linear factor $\alpha$ grows from 0 to 1 as training progresses, allowing the network to gradually adjust to the higher resolution as well as any new variables added. During the reinforcement stage, any unused layer for lower dimension are discarded as the grown network does more training. Our discriminator is trained progressively as well.

\paragraph{Encoder and UNet}

In an encoder-decoder structure using convolutional neural networks, the input is progressively down-sampled in the encoder and up-sampled in the decoder. For image translation task, some details and spatial information may be lost in the down-sampling process. UNet\cite{ronneberger2015u} is commonly used in image translation tasks\cite{isola2017image,zhang2017style} where details of the input image can be preserved in the output through the skip connection. We adapt such structure and our encoder mirrors the PGGAN generator structure -- growing as the generator grows to a higher resolution. The skip connection connects the encoding layers right before down-sampling with generator layers right after up-sampling. For details on the network structure please see \ref{tab:network_structure}.

\paragraph{Domain-adaptive Batch Renormalization}
Previous work such as \cite{dumoulin2017learned}, \cite{de2017modulating}, and \cite{miyato2018spectral} have shown that by using a different set of normalization parameters $(\gamma, \beta)$, one can train a generative network to output visually different images of the same object. Inspired by their discovery, we capture the style difference in the two domains $\mathbf{X_1}$ and $\mathbf{X_2}$ by using two sets of batch renormalization\cite{ioffe2017batch} parameters -- one for each domain. The motivation behind such design is as follows: Since both encoders are trying to encode the same semantic object represented in a different style, by sharing weights in all but the normalization layers, we encourage them to use the same latent encoding to represent the two visually different domains. Thus, different from prior works\cite{liu2017unsupervised,zhu2017unpaired,choi2017stargan,royer2017xgan} which share parameters only in the higher layers, we choose to share the weights for \textbf{all} layers except the batch renormalization layers and we use the same weight-sharing strategy for our two generators as well. This is the key to our TwinGAN model that enable us to capture shared semantic information and to train the network with fewer parameters. Due to the small batch size used at higher resolutions, we found empirically that batch renormalization performs better than batch normalization during inference. 

\subsection{Optimization}

We use three sets of losses to train our framework. The \textit{adversarial loss} ensures that the output images is indistinguishable from sampled images from the given domains. The \textit{reconstruction loss} enforces that the encoder-generator actually captures information in the input image and is able to reproduce the input. The \textit{cycle consistency} ensures that the input and output image contains the same features. 

\paragraph{Adversarial Loss}
Given Input domain $\mathbf{X_i} \in \{X_1, X_2\}$ and output domain $\mathbf{X_o} \in \{X_1, X_2\}$, the objective for image translation using vanilla GAN is formulated as

\[L_{VANILLA\_GAN_{i \rightarrow o}} = \mathbb{E}_{x_o \sim p_{x_o}(\mathbf{X_o})}[\log D_o(x_o)] + \mathbb{E}_{x_i \sim p_{x_i}(\mathbf{X_i})}[\log(1-D_o(G_o(E_i(x_i))))]\]

To boost the stability of GAN, we add the DRAGAN\cite{kodali2017convergence} objective function to our adversarial loss: 

\[L_{DRAGAN_{o}}=\lambda_{dragan} \cdot \mathbb{E}_{x_o \sim p_{x_o}, \delta \sim N_{d}(o, cI)} \|\Delta_{\mathbf{x}} D_{o}(x+\delta) \| - k \]

The adversarial objective is thus:

\[\min_{E,G}\max_{D}L_{GAN}(E,G,D) = \sum_{i\in \{1, 2\}}\sum_{o\in \{1, 2\}} L_{VANILLA\_GAN_{i \rightarrow o}} + L_{DRAGAN_{o}}\]

\paragraph{Cycle Consistency Loss}

We follow previous works\cite{zhu2017unpaired}\cite{royer2017xgan} and use a $L_1$ cycle consistency loss to encourage that the network acts as an identity operation for mapping images back to their own domains.

\begin{align*}
L_{cyc}(E,G) &= \mathop{x_1 \sim p_{x_1}(\mathbf{X_1})}(L_{1}(x_1, G_1(E_1(x_1)))) \\
               &+ \mathop{x_2 \sim p_{x_2}(\mathbf{X_2})}(L_{1}(x_2, G_2(E_2(x_2))))
\end{align*}

Optionally a discriminator can also be applied on the output $G_1(E_1(x_1)))$ and $G_2(E_2(x_2))$. Empirically we found that it leads to slightly better performance and less blurry output, but is by no means required.

\paragraph{Semantic Consistency Loss}
We would like the translated image to have the same semantic features as the input image. That is, the encoder should extract the same high level features for both input and output image regardless of their domains. Since cycle consistency already covers the semantic consistency for input and output from the same domain, here we focus on cross-domain semantic consistency. Similar to previous works \cite{royer2017xgan}, we formulate the semantic consistency loss as follows:
\begin{align*}
L_{sem}(E,G) &= \mathop{x_1 \sim p_{x_1}(\mathbf{X_1})}(L_{1}(E_1(x_1), E_2(G_2(E_1(x_1))))) \\
               &+ \mathop{x_2 \sim p_{x_2}(\mathbf{X_2})}(L_{1}(E_2(x_2), E_1(G_1(E_2(x_2)))))
\end{align*}

We note that it is generally not true that there exists a strictly one-to-one mapping between two domains. For example cat faces lack facial muscle to have as many expressions as we humans do, and forcing a one-to-one mapping between the two domains at the pixel level (e.g. using a loss such as $L_{1}(G_1(E_1(x_1)), G_1(E_2(G_2(E_1(x_1)))))$ in \cite{taigman2016unsupervised}) leads to mismatches. Therefore we choose to apply the loss only on the embeddings, which encode semantic information that is shared across domains. Thus, features unique to only one of the two domains are encouraged to be captured in the adaptive normalization parameters.

\paragraph{Overall Loss}
Our overall loss function is defined as: 
\[L_{total} = \lambda_{GAN} L_{GAN}  + \lambda_{cyc} L_{cyc} + \lambda_{sem} L_{sem}\]
Here the $\lambda$s are hyperparameters that control the weight on each of the objectives.

\begin{table}[h!]
  \begin{center}
    \begin{tabular}{l|c|c|c|c} % <-- Alignments: 1st column left, 2nd middle and 3rd right, with vertical lines in between
%     \textbf{Model} & \textbf{16} & \textbf{32} & \textbf{64} & \textbf{128} & \textbf{Average} \\
    \textbf{Source Image} & \includegraphics[width=1in]{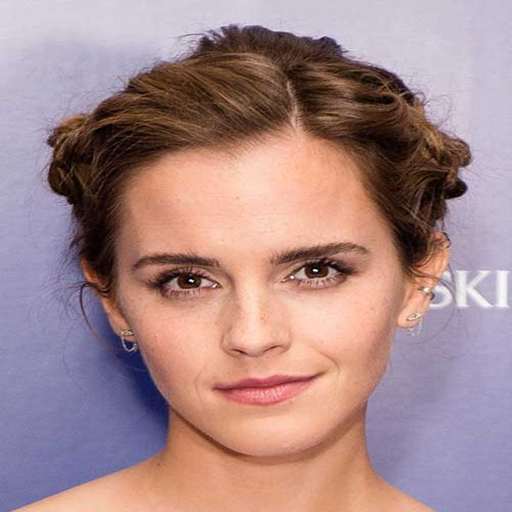} & \includegraphics[width=1in]{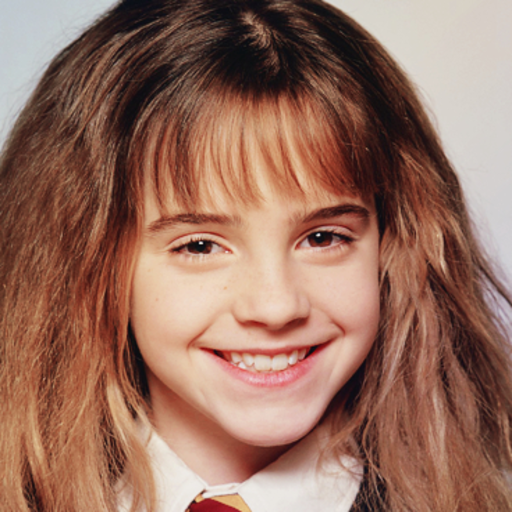} & \includegraphics[width=1in]{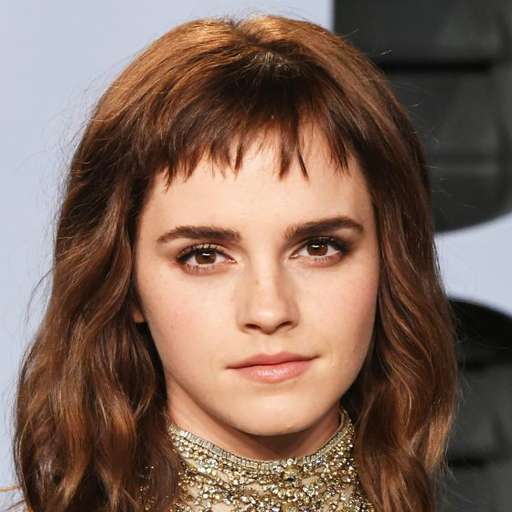} & \includegraphics[width=1in]{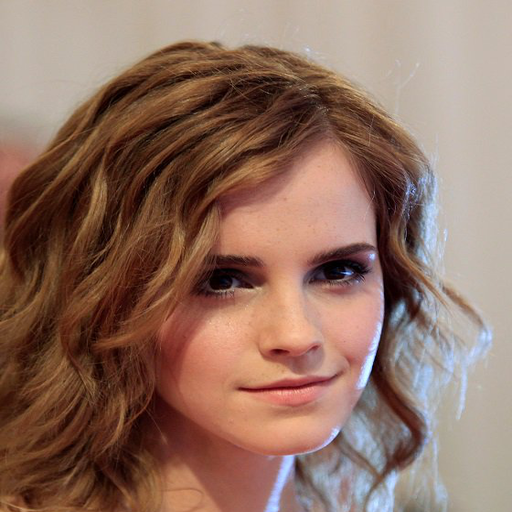}\\
    \hline
    \textbf{CycleGAN} & \includegraphics[width=1in]{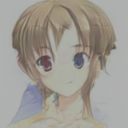} & \includegraphics[width=1in]{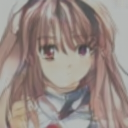} & \includegraphics[width=1in]{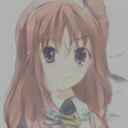} & \includegraphics[width=1in]{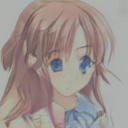}\\
    \textbf{UNIT} & \includegraphics[width=1in]{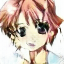} & \includegraphics[width=1in]{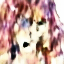} & \includegraphics[width=1in]{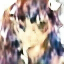} & \includegraphics[width=1in]{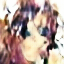}\\
    \textbf{MUNIT\textsubscript{1}} & \includegraphics[width=1in]{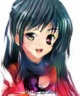} & \includegraphics[width=1in]{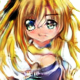} & \includegraphics[width=1in]{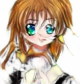} & \includegraphics[width=1in]{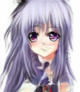}\\
    \textbf{MUNIT\textsubscript{2}} & \includegraphics[width=1in]{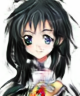} & \includegraphics[width=1in]{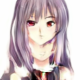} & \includegraphics[width=1in]{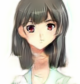} & \includegraphics[width=1in]{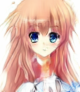}\\
    \textbf{TwinGAN(ours)} & \includegraphics[width=1in]{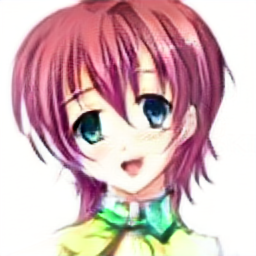} & \includegraphics[width=1in]{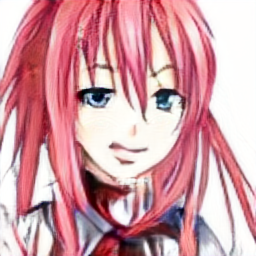} & \includegraphics[width=1in]{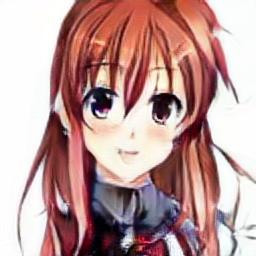} & \includegraphics[width=1in]{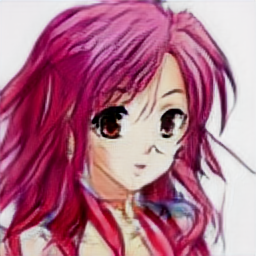}\\
    \end{tabular}
    \caption{Examples of unsupervised image translation from human to anime characters using various network structures. CycleGAN, UNIT, MUNIT are all trained to 64x64 resolution using the default settings from the official implementations. CycleGAN was trained for 465k iterations, UNIT for 940k iterations, and MUNIT for 650k iterations.  MUNIT\textsubscript{1} and MUNIT\textsubscript{2} uses the same network but different style embeddings. Notice that CycleGAN suffers from unable to move the eye position after translation, which looks unnatural on the target domain. UNIT failed to converge, while MUNIT suffers from inconsistent semantic correspondence between original and translated images.}
    \label{tab:table1}
  \end{center}
\end{table}

\section{Experiments}

We first compare our framework with other recent image translation architectures on the real-to-anime face translation task. We then study the benefits of cycle consistency loss and the semantic consistency loss. We show some use cases of a trained encoder and study its training process. Lastly, we show the generalization power of our framework on the human-to-cat face task.

\subsection{Real-to-Anime Face Translation} \label{sec:real_to_anime}

We train out network on two datasets: \textsc{CelebA} \cite{liu2015deep} dataset with 202599 celebrity face images and the "Getchu" dataset\cite{jin2017towards} containing 26752 anime character face images with clean background. During training, we randomly crop the images to $80\% - 100\%$ of their original size. In addition, we follow \cite{szegedy2017inception} and randomly flip the images and adjust the contrast, hue, brightness, and saturation levels. 

Regarding the hyperparameters in our framework, we set $\lambda_{GAN}=1, \lambda_{cyc}=1, \lambda_{sem}=0.1$(the hyperparameters are not fine-tuned due to limit on computing resources). Following \cite{karras2017progressive}, we use Adam\cite{kingma2014adam} with $\alpha= 0.001, \beta_1 = 0.5, \beta_2 = 0.99$, and $\epsilon = 10^{-8}$. We did not spend much effort on finding the optimal set of hyperparameters.

We start with a resolution of $4 \times 4$ and gradually grows to $256 \times 256$. We use a batch size of 8 for resolutions up to 64, and reduce that to 4 for 128 and 2 for 256 resolution. We show the discriminator 600k images for each stage of training. Different from \cite{karras2017progressive}, we find that DRAGAN has a shorter training time compared to variations of WGAN\cite{arjovsky2017wasserstein, gulrajani2017improved} and provides more stable training for image translation task. We used pixel-wise feature vector normalization but not equalized learning rate\cite{karras2017progressive}.

We compare our result with the same dataset trained with CycleGAN\cite{zhu2017unpaired}, UNIT\cite{liu2017unsupervised}, and the most recent MUNIT\cite{huang2018multimodal}. In \ref{tab:table1}, We found that our model outperforms all three in our experiments.
% TODO: maybe add a comparison on the network trained till 64x64 for fair comparison?

\subsection{Extra loss terms}

\begin{table}[h!]
  \begin{center}
    \begin{tabular}{l|c|c|c|c|c} % <-- Alignments: 1st column left, 2nd middle and 3rd right, with vertical lines in between
    \textbf{Model} & \textbf{16} & \textbf{32} & \textbf{64} & \textbf{128} & \textbf{Average} \\
    \hline
    No $L_{cyc}$ & 65.58 & 24.03 & 28.37 & 35.19 & 38.29\\
    No $L_{sem}$ & 40.15 & 11.60 & \textbf{9.75} & 12.01 & 18.38\\
    No UNet & 82.89 & 24.78 & 17.54 & 16.54 & 35.44\\
    With Style Embedding & \textbf{26.85} & 10.87 & 9.89 & 13.06 & 15.16\\
    No Style Embedding & 29.23 & \textbf{9.21} & 10.51 & \textbf{10.41} & \textbf{14.84}\\
    \end{tabular}
    \caption{Sliced Wasserstein Score}
    \label{tab:table2}
  \end{center}
\end{table}

We study the merits of the cycle loss, the semantic consistency loss, the style embedding, and the UNet. Because all requires extra computation during training, the extra training time spent must be justified by the better result they bring. We measure those benefits both quantitatively using Sliced Wasserstein Score proposed in \cite{karras2017progressive}.

In our experiments, we observed that having UNet encourages the network to find more local correspondences. Without UNet, the network failed to preserve correspondence between semantic parts and there were common error patterns such as the face direction becoming mirrored after translation -- which is technically allowed by all our loss terms but is judged as being unnatural by human. Note that similar error patterns are observed in our experiment with MUNIT \ref{tab:table1}, which does not use UNet.

Adding cycle loss and semantic consistency loss both resulted in higher Sliced Wasserstein Score and better output. Adding Style embedding increased the Sliced Wasserstein Score by a little, but it gave the user the ability to control some features such as hair color and eye color. However we argue that those features should perhaps belong to the content and the style embeddings failed to catch the more subtle yet important style information, such as eye-to-face ratio, texture of the hair, etc., that varies from painter to painter. We thus made the  style embedding optional and did not use that for the final results.

%  TODO: add section about using facenet embeddings.

\subsection{Human to cat face translation}

\begin{table}[h!]
  \begin{center}
    \begin{tabular}{l|c|c|c|c} % <-- Alignments: 1st column left, 2nd middle and 3rd right, with vertical lines in between
    \textbf{Source Image} & \includegraphics[width=1in]{original/IHIDUVI156JRG5N2NJGL6T0L6P76JWGU_0_square.png} & \includegraphics[width=1in]{original/IVW7BL5WMJRWF8TMXF9RNYE2O1VIPF13_0_square.png} & \includegraphics[width=1in]{original/K4NIGE15O9PB37WLK2JIL86R7V8RP0LB_0_square.png} & \includegraphics[width=1in]{original/KMJ2EAOPCPBXSH6P68S6IFH1TUVLY5E5_0_square.png}\\
    \hline
    \textbf{TwinGAN} & \includegraphics[width=1in]{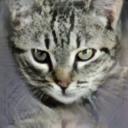} & \includegraphics[width=1in]{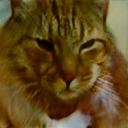} & \includegraphics[width=1in]{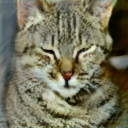} & \includegraphics[width=1in]{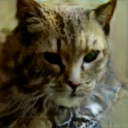}\\
    \end{tabular}
    \caption{Human to Cat Translation} %  and vice versa.
    \label{tab:table3}
  \end{center}
\end{table}

In order to show the general applicability of our model, here we show our results on the task of translating human faces to cat faces. For human face we collected 200k images from the \textsc{CelebA} dataset. We extracted around 10k cat faces from the CAT dataset\cite{zhang2008cat} by cropping the cat faces using the eye and ear positions\cite{alexia2018deep}. The network, the \textsc{CelebA} dataset, and training setup is the same as in \ref{sec:real_to_anime}.

\subsection{Learned Cross-domain Image Embeddings}

\begin{figure}[h!] \label{fig:nn_matching}
  \centering
    \includegraphics[width=1.0\textwidth]{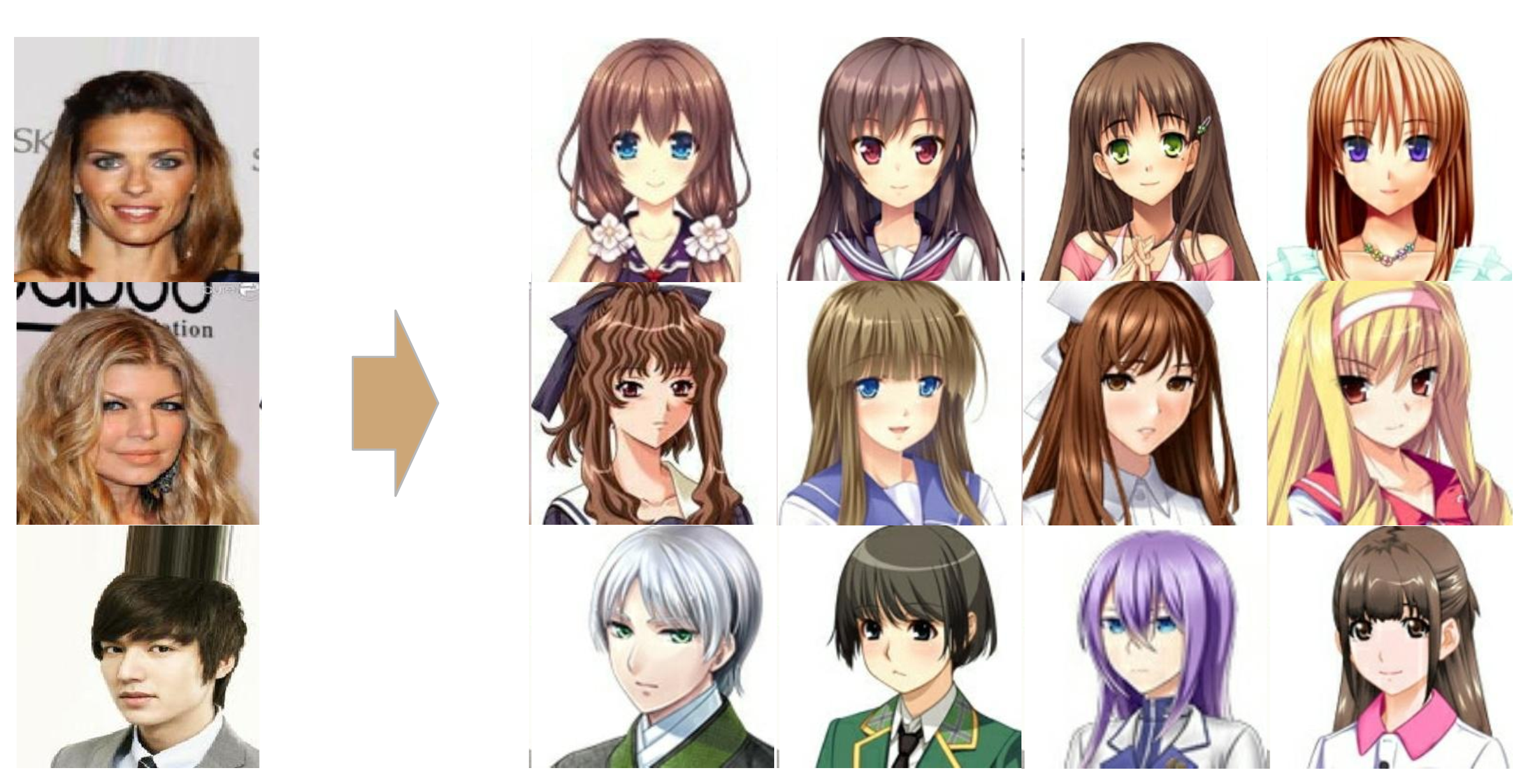}
  \caption{Human Portraits and their nearest neighbors in the Anime portraits domain, which are ranked from left to right the most to the least similar.}
\end{figure}

We want verify that meaningful semantic information shared across domains can indeed be extracted using our TwinGAN. For each domain, we use the corresponding encoder to extract the latent embeddings. For each image in domain $X_1$, we find the nearest neighbors in domain $X_2$ by calculating the cosine distances between the flattened embeddings. As shown in \ref{fig:nn_matching}, we found that meaningful correlations, including hair style, facing direction, sex, and clothing, can be established between the latent embeddings from the two domains.

\section{Conclusion and Future Work}
\label{conclusion}

We proposed the Twin-GAN framework that performs cross-domain image translation on unlabeled unpaired data. We demonstrated its use case on human-to-anime and human-to-cat face translation tasks. We showed that TwinGAN is capable of extracting common semantic information across the two domains while encoding the unique information in the adaptive normalization parameters.

Despite the success, there are still a lot of room for improvements. The process of selecting features to translate is not controllable in our current framework. Furthermore, when applied to even more diverse datasets involving changing point-of-view and reasoning about the 3d environment, our framework does poorly. We experimented with applying TwinGAN on image translation from the game of Minecraft to real street views -- our network collapsed at even low resolution. We hope to address these issues in future works.

\subsubsection*{Acknowledgments}

This work is not an official Google supported project and is developed solely by the author. We'd like to thank Alan Tian, Yanghua Jin, and Minjun Li for their inspirations. We'd like to thank all the creators of art works and fan-arts, without whom this research will not be possible.

% \section*{References}

\medskip
\bibliographystyle{plain}
\bibliography{unsupervised_domain_transfer.bib}

\newpage
\section{Supplimentary materials}
\subsection{Network Structure}
\begin{table}[!htb]\label{tab:network_structure}
  
    \begin{minipage}{0.5\linewidth}
 	  \centering
      \begin{tabular}{l|c|c}
      \textbf{Encoder} & Act. & Output shape \\
      \hline
      Input image & - & 3 x 256 x 256 \\
      Conv 1x1 & LReLU & 16 x 256 x 256 \\
      Conv 3x3 & LReLU & 16 x 256 x 256 \\
      Conv 3x3 & LReLU & 16 x 256 x 256 \\
      Downsample & - & 32 x 128 x 128 \\
      \hline
      Conv 3x3 & LReLU & 32 x 128 x 128 \\
      Conv 3x3 & LReLU & 64 x 128 x 128 \\
      Downsample & - & 64 x 64 x 64 \\
      \hline
      Conv 3x3 & LReLU & 64 x 64 x 64 \\
      Conv 3x3 & LReLU & 128 x 64 x 64 \\
      Downsample & - & 128 x 32 x 32 \\
      \hline
      Conv 3x3 & LReLU & 128 x 32 x 32 \\
      Conv 3x3 & LReLU & 256 x 32 x 32 \\
      Downsample & - & 256 x 16 x 16 \\
      \hline
      Conv 3x3 & LReLU & 256 x 16 x 16 \\
      Conv 3x3 & LReLU & 256 x 16 x 16 \\
      Downsample & - & 256 x 8 x 8 \\
      \hline
      Conv 3x3 & LReLU & 256 x 8 x 8 \\
      Conv 3x3 & LReLU & 256 x 8 x 8 \\
      Downsample & - & 256 x 4 x 4
      \end{tabular}
      \begin{tabular}{l|c|c}
      \textbf{Discriminator} & Act. & Output shape \\
      \hline
      Input image & - & 3 x 256 x 256 \\
      Encoder structure & - & 256 x 4 x 4 \\
      \hline
      Minibatch stddev & - & 257 x 4 x 4 \\
      Conv 3x3 & LReLU & 256 x 4 x 4 \\
      Conv 4x4 & LReLU & 256 x 1 x 1 \\
      Fully-connected & linear & 1 x 1 x 1
      \end{tabular}
	\end{minipage}\qquad
    \begin{minipage}{0.5\linewidth}
 	  \centering
      \begin{tabular}{l|c|c}
      \textbf{Generator} & Act. & Output shape \\
      \hline
      Latent Embedding & - & 256 x 4 x 4 \\
      Conv 3x3 & LReLU & 256 x 4 x 4 \\
      \hline
      Upsample & - & 256 x 8 x 8 \\
      Concat. UNet & - & 512 x 8 x 8 \\
      Conv 3x3 & LReLU & 256 x 8 x 8 \\
      Conv 3x3 & LReLU & 256 x 8 x 8 \\
      \hline
      Upsample & - & 256 x 16 x 16 \\
      Concat. UNet & - & 512 x 16 x 16 \\
      Conv 3x3 & LReLU & 256 x 16 x 16 \\
      Conv 3x3 & LReLU & 256 x 16 x 16 \\
      \hline
      Upsample & - & 256 x 32 x 32 \\
      Concat. UNet & - & 512 x 32 x 32 \\
      Conv 3x3 & LReLU & 128 x 32 x 32 \\
      Conv 3x3 & LReLU & 128 x 32 x 32 \\
      \hline
      Upsample & - & 128 x 64 x 64 \\
      Concat. UNet & - & 256 x 64 x 64 \\
      Conv 3x3 & LReLU & 64 x 64 x 64 \\
      Conv 3x3 & LReLU & 64 x 64 x 64 \\
      \hline
      Upsample & - & 64 x 128 x 128 \\
      Concat. UNet & - & 128 x 128 x 128 \\
      Conv 3x3 & LReLU & 32 x 128 x 128 \\
      Conv 3x3 & LReLU & 32 x 128 x 128 \\
      \hline
      Upsample & - & 32 x 256 x 256 \\
      Concat. UNet & - & 64 x 256 x 256 \\
      Conv 3x3 & LReLU & 16 x 256 x 256 \\
      Conv 3x3 & LReLU & 16 x 256 x 256 \\
      \hline
      Conv 1x1 & linear & 3 x 256 x 256
      \end{tabular}
	\end{minipage}
    \caption{TwinGAN Network Structure. Note that the discriminator borrows the same encoder structure but does not share weights with the encoder. In addition, a domain-adaptive batch renormalization layer is added after each convolution layer}
%   \end{center}
\end{table}

\end{document}